\documentclass[letterpaper, 10 pt, conference]{ieeeconf}  

\pdfoutput=1

\IEEEoverridecommandlockouts                              

\usepackage{graphics} 
\usepackage{graphicx}
\usepackage{float}
\usepackage{booktabs} 
\usepackage{array} 
\usepackage{caption} 
\usepackage{adjustbox}
\usepackage[utf8]{inputenc} 
\usepackage{balance}
\usepackage{hyperref}

\usepackage{tabularx} 
\newcolumntype{L}[1]{>{\raggedright\arraybackslash}p{#1}}
\newcolumntype{C}[1]{>{\centering\arraybackslash}p{#1}}

\usepackage[colorinlistoftodos]{todonotes}

\usepackage{geometry}
 \usepackage{amsmath,amssymb} 
\usepackage{color}

\let\labelindent\relax
\usepackage{enumitem}
\setlist[itemize]{nosep,     
                 topsep     = 0pt       ,
                 partopsep  = 0pt       ,
                 leftmargin = *         ,
                 after      = \vspace{-\baselineskip}
                 }

\usepackage{mdwlist}

\newcommand{\head}[1]{\textnormal{\textbf{#1}}}

\usepackage{mathrsfs} 

\def\etal{\emph{et al.}}

\usepackage{color}
\definecolor{markupcolour}{rgb}{0.7,0.0,0.1}

\graphicspath{{./img/}}

\title{\LARGE \bf
Recurrent Convolutional Fusion for RGB-D Object Recognition
}

\author{Mohammad Reza Loghmani$^{1}$, Mirco Planamente$^{2}$, Barbara Caputo$^{2}$ and Markus Vincze$^{1}$
\thanks{$^{1}$Mohammad Reza Loghmani and Markus Vincze are with the Vision4Robotics Group, Automation and Control Institute (ACIN), TU Wien, Vienna, Austria
        {\tt\small [loghmani, vincze]@acin.tuwien.ac.at}}%
\thanks{$^{2}$Barbara Caputo is with the VANDAL Laboratory, Italian Institute of Technology, Milan, Italy
        {\tt\small Mirco.Planamente, Barbara.Caputo]@iit.it}}%
}

\begin{document}

\bstctlcite{IEEEexample:BSTcontrol}


\maketitle
\thispagestyle{empty}
\pagestyle{empty}

\begin{abstract}

Providing robots with the ability to recognize objects like humans has always been one of the primary goals of robot vision. The introduction of RGB-D cameras has paved the way for a significant leap forward in this direction thanks to the rich information provided by these sensors. However, the robot vision community still lacks an effective method to synergically use the RGB and depth data to improve object recognition. In order to take a step in this direction, we introduce a novel end-to-end architecture for RGB-D object recognition called recurrent convolutional fusion (RCFusion). Our method generates compact and highly discriminative multi-modal features by combining RGB and depth information representing different levels of abstraction. Extensive experiments on two popular datasets, RGB-D Object Dataset and JHUIT-50, show that RCFusion significantly outperforms state-of-the-art approaches in both the object categorization and instance recognition tasks. In addition, experiments on the more challenging Object Clutter Indoor Dataset confirm the validity of our method in the presence of clutter and occlusion. The code is publicly available at: \hyperref[https://github.com/MRLoghmani/rcfusion]{``https://github.com/MRLoghmani/rcfusion''}.

\end{abstract}

\section{INTRODUCTION}
\label{sec:intro}

Human-built environments are, ultimately, collections of objects. Every daily activity requires to understand and operate a set of objects to accomplish a task. Robotic systems that aim at assisting the user in his own environment need to possess the ability to recognize objects. In fact, object recognition is the foundation for higher-level tasks that rely on an accurate description of the visual scene.

Despite the interesting results achieved for object recognition by operating on standard RGB images, there are inherent limitations due to the loss of data caused by projecting the 3-dimensional world into a 2-dimensional image plane. The use of RGB-D (Kinect-style) cameras potentially alleviates these shortcomings by using range imaging technologies to provide information about the camera-scene distance as a depth image. These sensors quickly became ubiquitous in robotics due to their affordable price and the rich visual information they provide. In fact, while the RGB image contains color, texture and appearance information, the depth image contains additional geometric information and is more robust with respect to lighting and color variations. Since RGB-D cameras are already deployed in most service robots, improving the performance of robot perceptual systems through a better integration of RGB and depth information would constitute a ``free lunch".

\begin{figure}[t]
\centering
\includegraphics[width=0.8\linewidth]{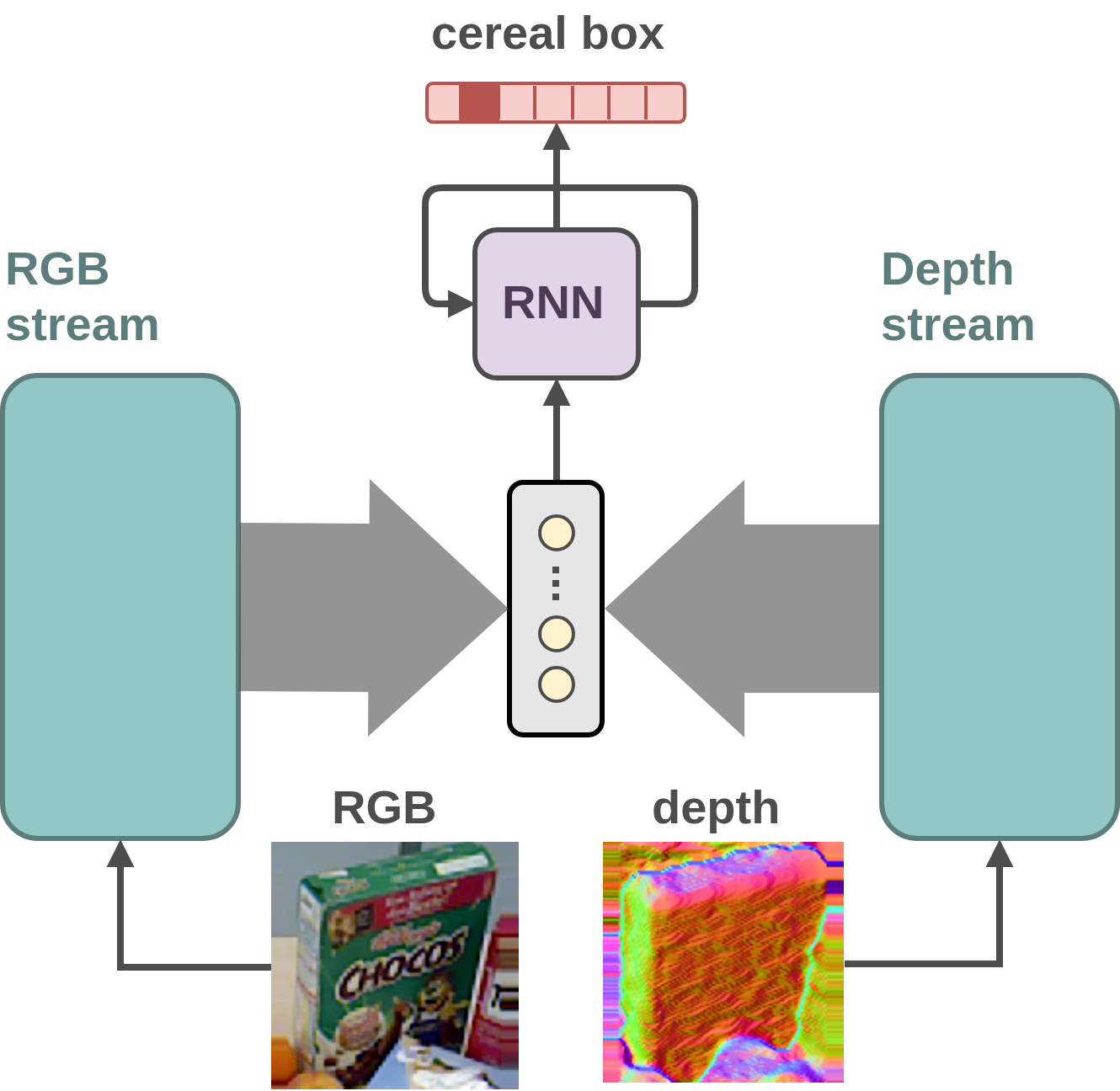}
\caption{High-level scheme of RCFusion. The blue boxes are instantiated with convolutional neural networks and the thick arrows represent multiple feature vectors extracted from different layers of a CNN.}
\label{fig:teaser}
\end{figure}

After the pivotal work of Krizhevsky~\etal~\cite{alexnet}, deep convolutional neural networks (CNNs) quickly became the dominant tool in computer vision, establishing new state-of-the-art results for a large variety of tasks, such as human pose estimation~\cite{openpose} and semantic segmentation~\cite{mask_rcnn}. Research in RGB-D object recognition followed the same trend, with numerous algorithms (e.g.~\cite{eitel2015,aakerberg2017,depthnet,deco,li2015,wang2015}) exploiting features learned from CNNs instead of the traditional hand-crafted features. The common pipeline involves two CNN streams, operating on RGB and depth images respectively, as feature extractors. However, the lack of a large-scale dataset of depth images to train the depth CNN forced the machine vision community to find practical workarounds. Much effort has been dedicated to develop methods that colorize the depth images in order to exploit CNNs pre-trained on RGB images. However, the actual strategies to extract and combine the features from the two modalities have been neglected. Several methods simply extract features from a specific layer of the two CNNs and combine them through a fully connected or a max pooling layer. We argue that these strategies are sub-optimal because (a) they are based on the assumption that the selected layer always represents the best abstraction level to combine RGB and depth information and (b) they do not exploit the full range of information from the two modalities during the fusion process. 

In this paper, we propose a novel end-to-end architecture for RGB-D object recognition called recurrent convolutional fusion (RCFusion). Our method extracts features from multiple hidden layers of the CNNs for RGB and depth, respectively, and combines them through a recurrent neural network (RNN), as shown in Figure~\ref{fig:teaser}. Our idea is that combining RGB and depth features from several levels of abstraction can provide greater information to the classifier to make the final prediction. We use RNNs to combine these features since they have been demonstrated as a very effective tool for selecting and retaining the most relevant information from a sequence of data in their memory unit~\cite{chung2014}. 

We evaluate our method on standard object recognition benchmarks, RGB-D Object Dataset~\cite{wrgbd} and JHUIT-50~\cite{jhuit50}, and we compare the results with the best performing methods in the literature. The experimental results show that our method outperforms the existing approaches and establishes new state-of-the-art results for both datasets. In order to further consolidate the effectiveness of our method, we adapt an object segmentation dataset, called Object Clutter Indoor Dataset (OCID) \cite{ocid}, to the instance recognition task to further evaluate RCFusion. OCID has been recently released to provide object scenes with high level of clutter and occlusion, arguably two of the biggest challenges faced by robotic visual perception systems~\cite{arid}. Our method confirms its efficacy also on this challenging dataset, despite the small amount of training data available. An implementation of the method, relying on tensorflow~\cite{tensorflow}, is publicly available at: \hyperref[https://github.com/MRLoghmani/rcfusion]{https://github.com/MRLoghmani/rcfusion}.\\In summary, our contributions are:
\begin{itemize}
    \item a novel architecture for RGB-D object recognition that sequentially combines RGB and depth features representing different levels of abstraction,
    \item state-of-the-art performance on the most popular RGB-D object recognition benchmark datasets,
    \item introduction of a new benchmark with robotic-oriented challenges, i.e. clutter, occlusion and few annotated samples per class.\\
\end{itemize}

The remainder of the paper is organized as follows: the next section positions our approach compared to related work, Section~\ref{sec:rcfusion} introduces the proposed method, Section~\ref{sec:exp} presents the experimental results and Section~\ref{sec:conclusion} draws the conclusions.


\section{RELATED WORK}
\label{sec:related_work}
The diffusion of RGB-D cameras fueled an increasing effort in designing visual algorithms able to exploit the additional depth information provided by these sensors. Classical approaches for RGB-D object recognition (e.g.~\cite{bo2011,wrgbd}) used a combination of different hand-crafted feature descriptors, such as SIFT, textons, and depth edges, on the two modalities (RGB and depth) to perform object matching. More recently, several methods have exploited shallow learning techniques to generate features from RGB-D data in an unsupervised learning framework~\cite{hmp2011,blum2012,socher2012}. 

Since the ground-breaking work of Krizhevsky~\etal~\cite{alexnet}, data-hungry deep CNNs have been the go-to solution for feature extraction. While large-scale datasets of RGB images, such as ImageNet~\cite{imagenet}, allowed the generation of powerful CNN-based models for RGB feature extraction, the lack of a depth counterpart posed the problem of how to extract features from depth images. An effective and convenient strategy to circumvent the problem is to colorize the depth images to exploit CNNs pre-trained on RGB data. Several hand-crafted colorization approaches have been proposed to map the raw depth value of each pixel~\cite{eitel2015} or derived physical quantities, such as position and orientation~\cite{gupta2014} or local surface normals~\cite{bo2013}, to colors. Carlucci~\etal~\cite{deco} proposed instead a leaning-based approach to colorize the depth images by training a colorization network. Other methods use alternatives to RGB-trained CNNs for extracting features from depth data. Li~\etal~\cite{li2015} generate the depth features using a modified version of HONV~\cite{honv} encoded with Fisher Vector~\cite{fisher}. Carlucci~\etal~\cite{depthnet} generate artificial depth data using 3D CAD models to train a CNN that extracts features directly from raw depth images. 

The aforementioned methods focus on effectively extracting features from the depth data and use trivial strategies to combine the two modalities for the final prediction. For example, Carlucci \etal~\cite{deco} simply select the class with the highest activation among the RGB and depth predictions, while Eitel \etal~\cite{eitel2015} and Aakerberg \etal~\cite{aakerberg2017} use a fully connected layer to learn to fuse the predictions from the two modalities. Alternatively, a few works prioritize the development of an effective modality fusion. Wang~\etal~\cite{wang2015} alternate between maximizing the discriminative characteristics of each modality and minimizing the inter-modality distance in feature space. Wang~\etal~\cite{cimdl} obtain the multi-modal feature by using a custom layer to separate the individual and correlated information of the extracted RGB and depth features. Both methods combine the two modalities by processing features extracted from one layer of the CNNs and rely on cumbersome multi-stage optimization processes.

Recent works from related areas, such as object detection and segmentation from color images, show the benefits of using features extracted from multiple layers of a CNN. Hariharan~\etal~\cite{hypercolumns}, increase the resolution of higher level features by combining information from lower layers at a pixel level for segmentation purposes. Bell~\etal~\cite{ionet} perform object detection at different scales using features extracted from different layers of a pre-trained network. These methods mostly take advantage of the difference in receptive fields in various layers of the neural network and use a simple combination of pooling and linear transformations to process the extracted features.  

The focus of this paper is on the synthesis of multi-modal features from RGB-D data rather than the depth processing. In fact, for the depth processing part, we adopt the well known colorization method based on surface normals, since it has been proved to be effective and robust across tasks~\cite{bo2013,cimdl,aakerberg2017}. Differently from existing works, our method produces highly informative global features from different levels of abstraction through a dedicated non-linear unit, called projection block. Features from the RGB and depth modalities are then combined together in a sequential manner to generate the final multi-modal representation. In addition, our model can be trained end-to-end, without the need of optimizing in multiple stages.

\section{RECURRENT CONVOLUTIONAL FUSION}
\label{sec:rcfusion}

\begin{figure*}[t]
\centering
\includegraphics[width=0.65\linewidth]{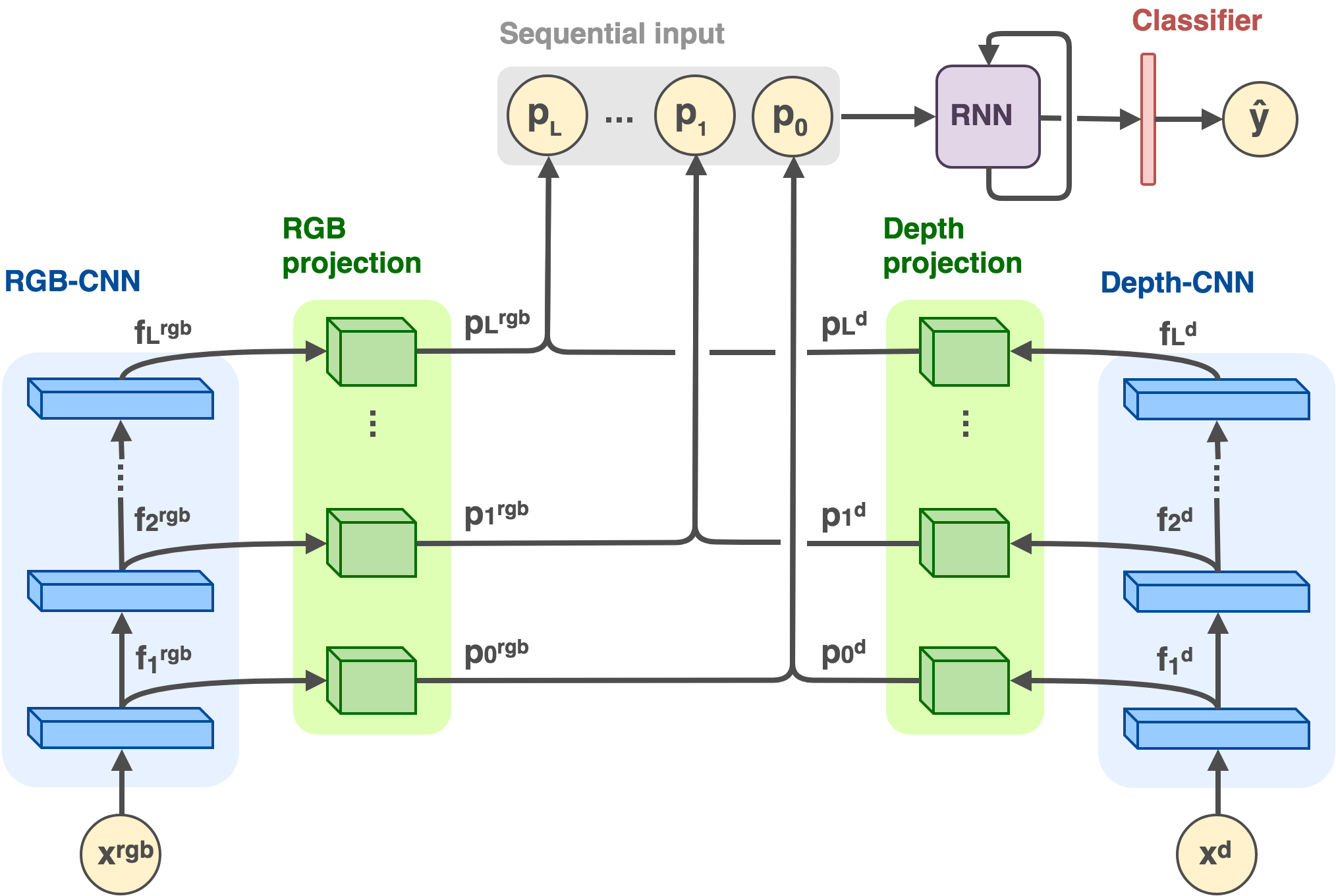}
\caption{Architecture of recurrent convolutional fusion. It consists of two streams of convolutional neural networks (CNN) that process RGB and depth images, respectively. The output of corresponding hidden layers from the two streams are projected into a common space, concatenated and sequentially fed into a recurrent neural network (RNN) that synthesizes the final multi-modal features. The output of the RNN is then used by a classifier to determine the label of the input data.}
\label{fig:rcfusion}
\end{figure*}

Our multi-modal deep neural network for RGB-D object recognition is illustrated in Figure~\ref{fig:rcfusion}. The network's architecture has three main stages: 
\begin{enumerate}
\item \textit{multi-level feature extraction:} 
two streams of convolutional networks, with the same architecture, are used to process RGB and depth data (RGB-CNN and Depth-CNN), respectively, and extract features at different levels of the networks;
\item \textit{feature projection and concatenation: }
features extracted from each level of the RGB- and Depth-CNN are individually transformed through projection blocks and concatenated to create the corresponding RGB-D feature;
\item \textit{recurrent multi-modal fusion:} 
RGB-D features extracted from different levels are sequentially fed to an RNN that produces a descriptive and compact multi-modal feature.
\end{enumerate}
The output of the recurrent network is then used by a softmax classifier to infer the object label. The network can be trained end-to-end with a cross-entropy loss using standard backpropagation algorithms based on stochastic gradient descent. In the following, we describe in greater detail each of the aforestated characteristics of RCFusion.

\subsection{Multi-level Feature Extraction}

CNNs process the input with sets of filters learned from a large amount of data. These filters represent progressively higher levels of abstraction, going from the input to the output: edges, textures, patterns, parts, and objects~\cite{zeiler2014, olah2017}. Methods for RGB-D object recognition commonly combine the output of one of the last layers of the RGB- and Depth-CNN (typically the last layer before the classifier) to obtain the final multi-modal feature. This strategy is based on the strong assumption that the chosen layer represents the appropriate level of abstraction to combine the two modalities. We argue that it is possible to remove this assumption by combining RGB and depth information at multiple layers across the CNNs and use them all to generate a highly discriminative RGB-D feature.
Let us denote with $x^{rgb} \in \mathcal{X}^{rgb}$ the RGB input images, with $x^{d} \in \mathcal{X}^{d}$ the depth input images and $y \in \mathcal{Y}$ the labels, where $\mathcal{X}^{rgb}$, $\mathcal{X}^{d}$ and $\mathcal{Y}$ are the RGB/depth input and label space. We further denote with $f^{rgb}_i$ and $f^{d}_i$ the output of layer $i$ of RGB-CNN and Depth-CNN, respectively, with $i=1,...,L$ and $L$ the total number of layers of each CNN. Notably, visualizing the learned filters has shown ~\cite{zeiler2014,olah2017} that, for a given task, a chosen CNN architecture consistently generates features with the same level of abstraction from a reference layer. For example, AlexNet~\cite{alexnet} learns various types of Gabor filters in the first convolutional layer. So, the same architecture is chosen for RGB- and Depth-CNN to ensure the same abstraction level at corresponding layers.

\begin{figure}[t]
\centering
\includegraphics[width=1.0\linewidth]{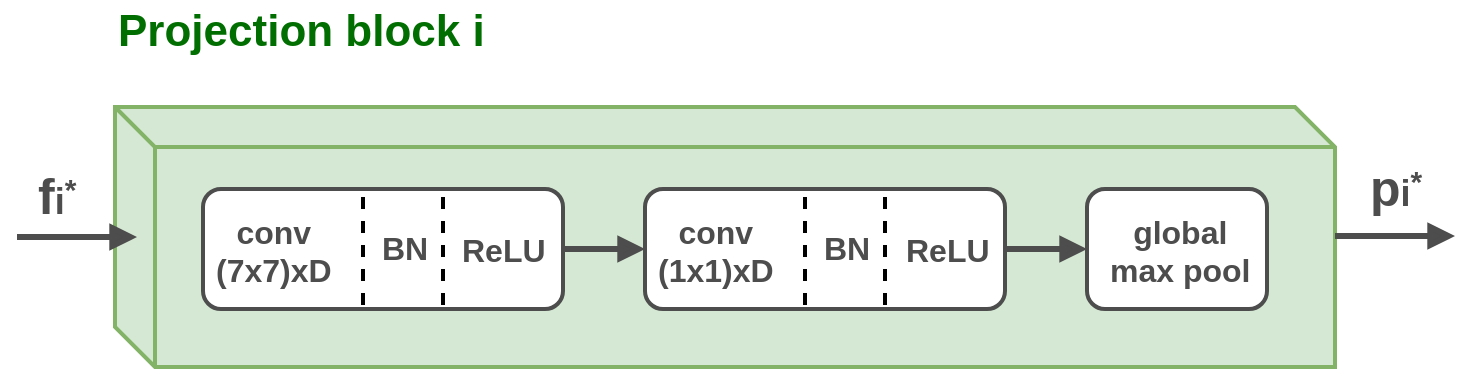}
\caption{Implementation of the projection block that transforms the feature $f^{*}_i$ into the projected feature $p^{*}_i$. $conv(k\times k)\times D$ indicates a convolutional layer with $D$ filters of size $(k\times k)$, $BN$ indicates a batch normalization layer and $ReLU$ indicates an activation layer with ReLU non-linearity.}
\label{fig:proj_block}
\end{figure}

\subsection{Feature Projection and Concatenation}
\label{subsec:projection}

One of the main challenges in combining features obtained from different hidden layers of the same network is the lack of a one-to-one correspondence between elements of the different feature vectors. More formally, $f^{*}_i$ and $f^{*}_j$, with $i \neq j$ and $*$ indicating any of the superscripts $rgb$ or $d$, have (in general) different dimensions and thus belong to distinct feature spaces, $\mathcal{F}_i$ and $\mathcal{F}_j$. In order to make features coming from different levels of abstraction comparable, we project them into a common space $\bar{\mathcal{F}}$:
\begin{equation}
\setlength\abovedisplayskip{5pt}
\setlength\belowdisplayskip{5pt}
  p^{*}_i = G^{*}_i(f^{*}_i) \quad \text{s.t.} 
  \quad p^{*}_i \in \bar{\mathcal{F}} 
  \label{eq:risk}
\end{equation}
The projection block $G_i(.)$ performs a set of non-linear operations to transform a volumetric input into a vector of dimensions (1$\times$D). More specifically, $G_i(.)$ is defined by two convolutional layers (with batch normalization and ReLU non-linearity) and a global max pooling layer, as shown in Figure \ref{fig:proj_block}. The projected RGB and depth features of each layer $i$ are then concatenated to form $p_i = \big[ p^{rgb}_i ; p^{d}_i \big]$.

\subsection{Recurrent Multi-modal Fusion}
\label{subsec:fusion}

In order to create a compact multi-modal representation, the set $\big\{ p_1, \dots , p_{L} \big\}$ is sequentially fed to an RNN. Recurrent models align the positions of the elements in the sequence to steps in computation time and generate a sequence of hidden states $h_i$ as a function of the previous hidden state $h_{i-1}$ and the current input $p_i$. In this paper, we use an instantiation of an RNN called gated recurrent unit (GRU) \cite{gru}. This network is considered to be a variation of long-short term memory (LSTM)~\cite{lstm} and its effectiveness in dealing with long input sequences has been repeatedly shown~\cite{chung2014}. Despite performing comparably, GRU requires $25\%$ fewer parameters than LSTM. 

GRU computes the $n^{th}$ element of the hidden state at step $i$ as an adaptive linear interpolation:
\begin{equation}
\setlength\abovedisplayskip{5pt}
\setlength\belowdisplayskip{5pt}
  h_i^n = (1-z_i^n )h_{i-1}^n + z_i^n \tilde{h}_i^n ,
  \label{eq:h}
\end{equation}
where $z_i^n$ is called update gate and is computed as 
\begin{equation}
\setlength\abovedisplayskip{5pt}
\setlength\belowdisplayskip{5pt}
  z_i^n = sigmoid(\theta_z p_i + \gamma_z h_{i})^n,
  \label{eq:z}
\end{equation}
where $sigmoid(.)$ is the sigmoid function and $\theta_z$ and $\gamma_z$ are the trainable parameters of the gate. Essentially, the update gate determines how much the unit updates its content. The candidate activation $\tilde{h}_i$ in Equation \ref{eq:h} is computed as
\begin{equation}
\setlength\abovedisplayskip{5pt}
\setlength\belowdisplayskip{5pt}
  \tilde{h}_i^n = tanh(\theta_h p_i + \gamma_h (r_i \odot h_{i-1}))^n ,
  \label{eq:h_tilde}
\end{equation}
where $r_i$ is the reset gate, $\theta_h$ and $\gamma_h$ are trainable parameters and $\odot$ is the element-wise multiplication operation. Similarly to $z_i^n$, the reset gate $r_i^n$  is computed as 
\begin{equation}
\setlength\abovedisplayskip{5pt}
\setlength\belowdisplayskip{5pt}
  r_i^n = sigmoid(\theta_r p_i + \gamma_r h_{i})^n,
  \label{eq:r}
\end{equation}
where $\theta_r$ and $\gamma_r$ are the trainable parameters of the gate. When $r_i^n$ assumes values close to zero, it effectively resets the hidden state of the network to the current input $p_i$. This double-gate mechanism has the goal of ensuring that the hidden state progressively embeds the most relevant information of the input sequence $\big\{ p_1, \dots , p_{L} \big\}$.

The RNN, combined with a softmax classifier, models a probability distribution over a sequence by being trained to predict the category label given the sequence of projected RGB-D features. In particular, the prediction of the $j^{th}$ class of the multinomial distribution of $K$ object categories is obtained as 
\begin{equation}
\hat{y}_j = Pr(y_j=1 | p_1, ..., p_1)=\frac{exp(h^{T}_L\theta_c^j )}{\sum_{k=1}^{K}exp(h^{T}_{L}\theta_c^k)},
  \label{eq:pred}
\end{equation}
where $\theta$ is the matrix of trainable parameters of the classifier and $\theta_{j(/_k)}$ represents its $j^{th}(/k^{th})$ row, and the superscript $T$ represents the transpose operation.

The choice of a recurrent network for this operation is twofold: (a) the hidden state of the network acts as a memory unit and embeds a summary of the most relevant information from the different levels of abstraction, and (b) the number of parameters of the network is independent of $L$, while for a more straightforward choice, such as a fully connected layer, it grows linearly with $L$. Although RNNs are typically used to process time series data, our atypical deployment is supported by previous works ~\cite{schmidhuber1996,mahoney2000} that have shown that these type of networks are also useful in compressing and combining information from different sources. Our intuition, validated by experiments in Section~\ref{sec:exp}, is that a recurrent network can be effectively used to aggregate the RGB and depth features from different levels of abstraction. 

\section{EXPERIMENTS}
\label{sec:exp}

In the following, we evaluate RCFusion on RGB-D Object Dataset, JHUIT-50, and OCID. After revealing the protocol used to set up the experiments, we discuss the setting used for training the network. Then, we show how the performances of our method compare to the existing literature. Finally, we perform an ablation study to identify the contribution of the different elements of our method.

\subsection{Datasets}
\label{subsec:datasets}


\textbf{RGB-D Object Dataset:} It contains 41,877 RGB-D images capturing 300 objects from 51 categories, spanning from fruit and vegetables to tools and containers. Since its introduction, this dataset has become the silver thread connecting the existing methods for RGB-D object recognition. We use this dataset to assess the performance of RCFusion in the object categorization task. For the evaluation, we follow the standard experimental protocol defined in~\cite{bo2011}, where ten training/test split are defined in such a way that one object instance per class is left out of the training set. The reported results are the average accuracy over the different splits. 

\textbf{JHUIT-50:} It contains 14,698 RGB-D images capturing 50 common workshop tools, such as clamps and screw drivers. Since this dataset presents few objects, but very similar to each other, it can be used to assess the performance of RCFusion in the instance recognition task. For the evaluation, we follow the standard experimental protocol defined in~\cite{jhuit50}, where training data are collected from fixed viewing angles between the camera and the object while the test data are collected by freely moving the camera around the object. 

\textbf{OCID:} It includes 96 cluttered scenes representing common objects organized in three subsets: ARID20, ARID10, and YCB10. The ARID20 and ARID10 subsets contain scenes that include, respectively, up to 20 and 10 out of 59 objects from Autonomous Robot Indoor Dataset \cite{arid}. Similarly, the YCB10 subset contains scenes with up to 10 objects from Yale-CMU-Berkeley object and model set~\cite{ycb}. Each scene is built incrementally by adding one object at a time and recording new frames at each step. Two ASUS-PRO cameras, positioned at different heights, are used to simultaneously record each scene. Further scene variation is introduced by changing the support plane (floor and table) and the background texture. Since OCID has been acquired to evaluate object segmentation methods in cluttered scenes, semantic labels are not provided by the authors. In order to adapt this dataset to a classification task, we crop out the objects from each frame and annotate them with semantic labels similar to the RGB-D Object Dataset. To avoid redundancies, we go sequentially through the frames of each scene and save only the crops that have an overlap with the bounding box of a newly introduced object. We then filter out the classes with less than 20 images to ensure a minimum amout of training samples per class. We use the crops from the ARID20 subset to train the network for an instance recognition task and then use the crops from the ARID10 subset for testing. Overall, we obtain 3,939 RGB-D images capturing 49 distinct objects. The original datasets, as well as the crops and annotation used in this paper are available at \hyperref[https://www.acin.tuwien.ac.at/en/vision-for-robotics/software-tools/object-clutter-indoor-dataset/]{``https://www.acin.tuwien.ac.at/en/vision-for-robotics/software-tools/object-clutter-indoor-dataset/''}. 

\subsection{Architecture}
The network architecture of RCFusion passes through independent design choices of three main elements: RGB-/Depth-CNN, projection blocks and RNN.

\textbf{RGB-/Depth-CNN: } With computational and memory efficiency in mind, we choose a CNN architecture with a relatively small number of parameters. Since residual networks have become a standard choice, we deploy ResNet-18, the most compact representation proposed by He~\etal~\cite{resnet}. ResNet-18 has 18 convolutional layers organized in five residual blocks, for a total of approximately 40,000 parameters. We extract features from two layers of each residual block to pass to the projection blocks. An implementation of ResNet-18 pre-trained on ImageNet is available in~\cite{resnet-github}.

\textbf{Projection blocks: } The projection blocks, shown in Figure \ref{fig:proj_block}, are designed in such a way that the first convolutional layer focuses on exploiting the spatial dimensions, width and height, of the input with $D$ filters of size (7$\times$7), while the second convolutional layer exploits its depth with $D$ filters of size (1$\times$1). Finally, the global max pooling computes the maximum of each depth slice. This instantiation of the projection blocks has provided the best performances among those that we tried.

\textbf{RNN: } In a trade-off between network capacity and small number of parameters, we use the popular GRU~\cite{gru}. In our experiments, we process the sequence of projected vectors with a single GRU layer with a number of memory neurons $M$. An implementation of GRU can be found in all the most popular deep learning libraries, including tensorflow.

\subsection{Training}
\label{subsec:training}
We train our model using RMSprop optimizer with batch size $64$, learning rate $0.001$, momentum $0.9$, weight decay $0.0002$ and max norm $4$. The architecture specific parameters have been fixed through a grid search to projection depth $D=512$ and memory neurons $M=50$. The weights of the two ResNet-18 used as the RGB- and Depth-CNN are initialized with values obtained by pre-training the networks on ImageNet. The rest of the network is initialized with Xavier initialization method in a multi-start fashion, where the network is initialized multiple times and, after one epoch, only the most promising model continues the training. All the parameters of the network, including those defining the RGB- and Depth-CNN, are updated during training. The input to the network is synchronized RGB and depth images pre-processed following the procedure in~\cite{aakerberg2017}, where the depth information is encoded with surface normals (see Section \ref{sec:related_work}). For JHUIT-50 and OCID, we compensate for the small training set with simple data augmentation techniques: scaling, horizontal and vertical flip, and $90$ degree rotation.

\subsection{Results}
\label{subsec:results}

In order to validate our method, we first compare the performance of RCFusion to existing methods on two benchmark datasets, RGB-D Object Dataet and JHUIT-50. We then test our method on a more challenging dataset, OCID, and perform an ablation study to showcase the contribution of each component of the method.

\begin{figure*}[t]
\centering
\includegraphics[width=0.95\textwidth]{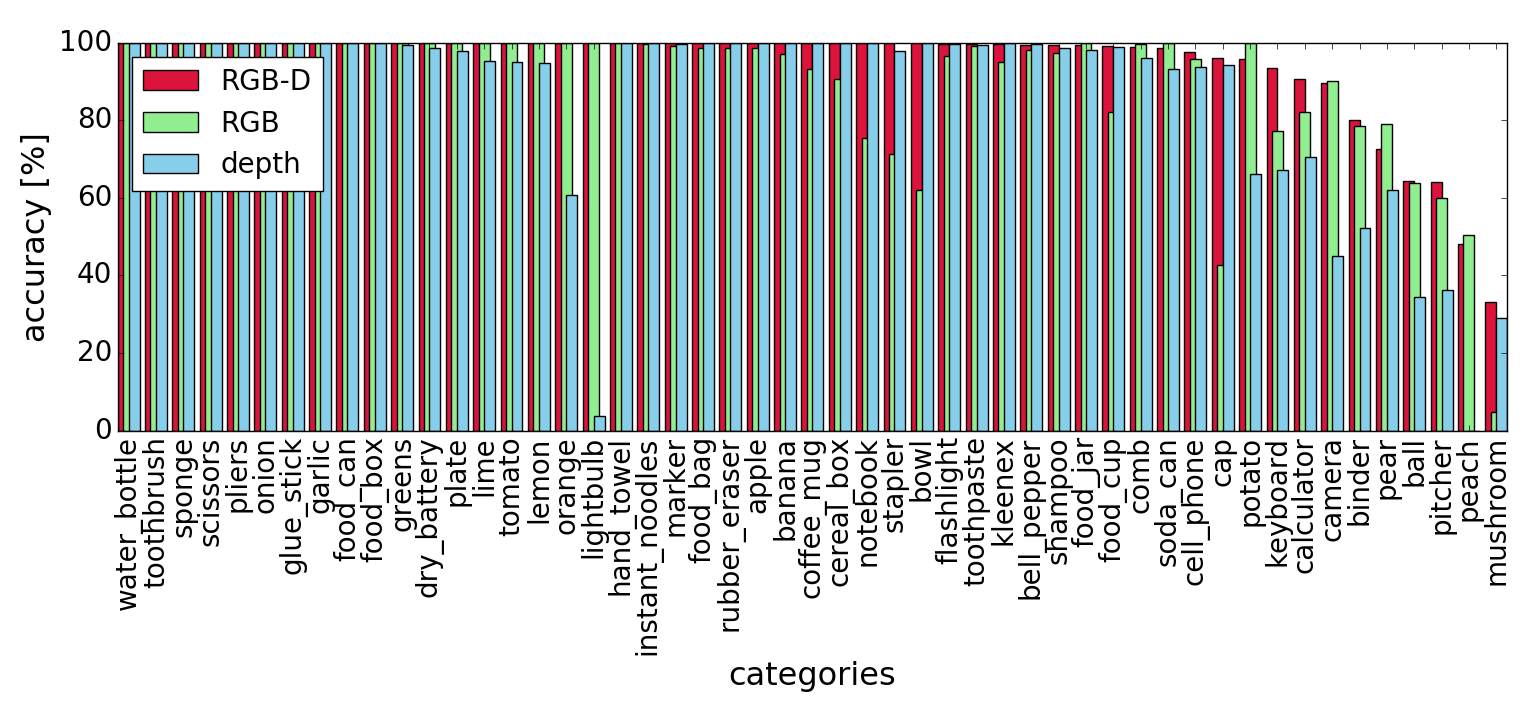}
\caption{Per class accuracy ($\%$) of RCFusion on RGB-D Object Dataset~\cite{wrgbd}.}
\label{fig:per_class}
\end{figure*}

\setlength{\tabcolsep}{7pt}
\begin{table}[t]
\begin{center}
\caption{Accuracy (\%) of several methods for object recognition on RGB-D Object Dataset~\cite{wrgbd}. Red: highest result; blue: other considerable results.}
\label{tab:wrgbd}
\begin{tabular}{lccc}
\hline\noalign{\smallskip}
\multicolumn{4}{c}{\head{RGB-D Object Dataset}} \\
\noalign{\smallskip}
\cline{1-4}
\noalign{\smallskip}
Method & RGB & Depth & RGB-D \\
\noalign{\smallskip}
\hline
\noalign{\smallskip}
LMMMDL~\cite{wang2015} &
 74.6$\pm$2.9 & 75.5.8$\pm$2.7 & 86.9$\pm$2.6\\
FusionNet~\cite{eitel2015} &
 84.1$\pm$2.7 & 83.8$\pm$2.7 & 91.3$\pm$1.4 \\
 CNN w/ FV~\cite{li2015} &
 {\color{red}\textbf{90.8}}$\pm$1.6& 81.8$\pm$2.4 & {\color{blue}\textbf{93.8}}$\pm$0.9 \\
 DepthNet~\cite{depthnet} &
 88.4$\pm$1.8 & 83.8$\pm$2.0 & 92.2$\pm$1.3 \\
 CIMDL~\cite{cimdl} &
 87.3$\pm$1.6 &  {\color{blue}\textbf{84.2}}$\pm$1.7 & 92.4$\pm$1.8 \\
 FusionNet enhenced~\cite{aakerberg2017} &
 {\color{blue}\textbf{89.5}}$\pm$1.9 & {\color{blue}\textbf{84.5}}$\pm$2.9 & {\color{blue}\textbf{93.5}}$\pm$1.1 \\
 DECO~\cite{deco} &
 {\color{blue}\textbf{89.5}}$\pm$1.6 & 84.0$\pm$2.3 & {\color{blue}\textbf{93.6}}$\pm$0.9 \\
  \textbf{RCFusion} &
 {\color{blue}\textbf{89.6}}$\pm$2.2 & {\color{red}\textbf{85.9}}$\pm$2.7 & {\color{red}\textbf{94.4}}$\pm$1.4 \\
\noalign{\smallskip}
\hline
\end{tabular}
\end{center}
\end{table}
\setlength{\tabcolsep}{1.4pt}

\setlength{\tabcolsep}{12pt}
\begin{table}[t]
\begin{center}
\caption{Accuracy (\%) of several methods for object recognition on JHUIT-50~\cite{jhuit50}. Red: highest result; blue: other considerable results.}
\label{tab:jhuit}
\begin{tabular}{lccc}
\hline\noalign{\smallskip}
\multicolumn{4}{c}{\head{JHUIT-50}} \\
\noalign{\smallskip}
\cline{1-4}
\noalign{\smallskip}
Method & RGB & Depth & RGB-D \\
\hline\noalign{\smallskip}
 DepthNet~\cite{depthnet} &
 88.0 & 55.0 & 90.3 \\
 FusionNet enhanced~\cite{aakerberg2017} &
 {\color{blue}\textbf{94.7}} & 56.0 & {\color{blue}\textbf{95.3}} \\
 DECO~\cite{deco} &
 {\color{blue}\textbf{94.7}} & {\color{red}\textbf{61.8}} & {\color{blue}\textbf{95.7}} \\
 \textbf{RCFusion} &
 {\color{red}\textbf{95.1}} & {\color{blue}\textbf{59.8}} & {\color{red}\textbf{97.7}} \\
\noalign{\smallskip}
\hline
\end{tabular}
\end{center}
\end{table}
\setlength{\tabcolsep}{1.4pt}


\setlength{\tabcolsep}{11pt}
\begin{table}[t]
\begin{center}
\caption{Most frequently misclassified classes in RGB, depth and RGB-D for selected reference classes. }
\label{tab:confusion}
\begin{tabular}{l|ccc}
\hline\noalign{\smallskip}
\multicolumn{4}{c}{\head{Misclassification cases}} \\
\noalign{\smallskip}
\cline{1-4}
\noalign{\smallskip}
Reference class & RGB & Depth & RGB-D \\
\hline\noalign{\smallskip}
 calculator &
 keyboard & hand towel & hand towel \\
 keyboard &
 calculator & binder & calculator \\
 pear &
 apple & apple & apple \\
 potato &
 lime & lime & lime \\
\noalign{\smallskip}
\hline
\end{tabular}
\end{center}
\end{table}
\setlength{\tabcolsep}{1.4pt}

\textbf{Benchmark: } We benchmark RCFusion on RGB-D Object Dataset and JHUIT-50 against other methods in the literature. Table~\ref{tab:wrgbd} shows the results on RGB-D Object Dataset for the object categorization task. The reported results for the RGB and depth modality are obtained by training a classifier on the final features of the RGB- and Depth-CNN, respectively. The reported multi-modal RGB-D results show that our method outperforms all the competing approaches. In addition, the results of the single modality predictions demonstrate that ResNet-18 is a valid trade-off between small number of parameters and high accuracy. In fact, on the RGB modality, the accuracy is second only to~\cite{li2015}, where they use a VGG network~\cite{vgg} that introduces considerably more parameters than ResNet-18. For the depth modality, ResNet-18 provides higher accuracy than all the competing methods. 

In order to gain a better insight on the performance of RCFusion, we consider the accuracy on the individual categories of RGB-D Object Dataset. Figure~\ref{fig:per_class} shows that the multi-modal approach either matches or improves over the results on the single modalities for almost all categories. For categories like ``lightbulb", ``orange" or ``bowl", where the accuracy on one modality is very low, RCFusion learns to rely on the other modality. An interesting insight on the functioning of the method is given by comparing, for each category, which other categories generate the misclassification. Table \ref{tab:confusion} indicates, for few example classes, the most frequently misclassified class in the RGB, depth and RGB-D case\footnote{Due to the relatively large number of classes, it was not possible to show the confusion matrices of the three modalities within a reasonable space.}. When an object class is confused with distinct classes in the individual modalities, like for ``keyboard" and ``calculator", the RGB-D modality can perform better. However, when an object class is confused with the same classes in both RGB and depth modalities, like for ``pear" and ``potato", the RGB-D modality can perform slightly worse than the single modalities. This highlights a weakness of the method that will be the subject of future investigations.

Table~\ref{tab:jhuit} shows the results on JHUIT-50 for the instance recognition task. For the individual modalities, ResNet-18 shows again a compelling performance. In the multi-modal RGB-D classification, our method clearly outperforms all the competing approaches with a margin of $2\%$ on the best existing method, DECO~\cite{deco}. In summary, RCFusion establishes new state-of-the-art results on the two most popular datasets for RGB-D object recognition, demonstrating its robustness against changes in the dataset and the task.

\begin{figure}[h]
\centering
\includegraphics[width=0.9\linewidth]{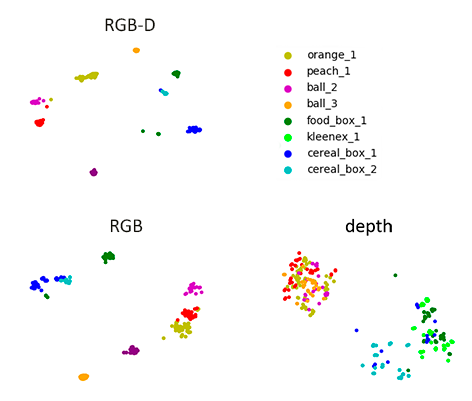}
\caption{t-SNE visualization of the final features obtained for RGB, depth and RGB-D modalities.}
\label{fig:tsne}
\end{figure}

\begin{figure}[t]
\centering
\includegraphics[width=0.8\linewidth]{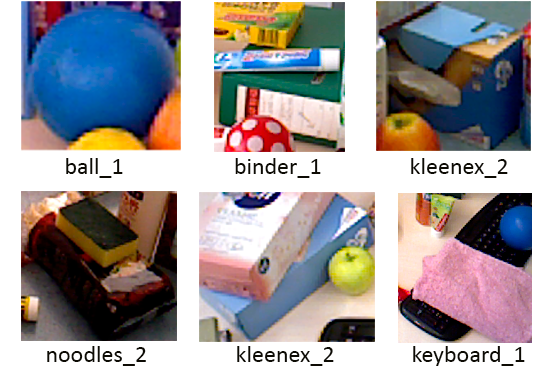}
\caption{Examples of object crops from the Object Cluttered Indoor Dataset~\cite{ocid} with their instance label.}
\label{fig:ocid_examples}
\end{figure}

\textbf{Challenge: } In order to evaluate the performance of our method on more robotic-oriented data, we show experiments on OCID. This dataset has been recorded with the specific goal of creating highly cluttered and occluded object scenes. Some example images are shown in Figure \ref{fig:ocid_examples}. In addition, its small training set of 2,428 cropped images represents an additional challenge. Table \ref{tab:ocid} shows the results on OCID for the instance recognition task. As well as our method, we also report the results of DECO, that showed competitive performance on RBG-D Object Dataset and JHUIT-50. The results on the single modalities show that the depth data alone are not very informative for this task, with a gap of $~50\%$ with respect to the RGB modality. Despite this difference, our method leverages both modalities and obtains an improvement of $6.1\%$ in accuracy with respect to the RGB modality alone. On the contrary, DECO reveals its limits and maintains the same performance of the RGB modality even in the multi-modal case. This result is due to the simple strategy used in DECO for the multi-modal fusion: the final prediction is made by simply taking the class with the maximum probability among the RGB and depth predictions. The more complex modality fusion of RCFusion thus translates into over $10\%$ improvement in accuracy with respect to DECO.

\textbf{Feature analysis:} An interesting intuition of the effectiveness of RCFusion comes from the visualization of the features learned on the OCID dataset. Figure~\ref{fig:tsne} represents the two dimensional t-SNE embedding of the final features of the different modalities. As expected, the t-SNE embedding of the depth features clusters together objects with similar shapes. For example, objects with near-spherical shapes like ``orange\_1", ``pear\_1" and ``ball\_2(/3)" are grouped together. The RGB modality provides more discriminative features, but similar pairs of objects, like (``orange\_1"-``peach\_1") and (``cereal\_box\_1"-``cereal\_box\_2") are very close to each other. Instead, the embedding of the RGB-D features neatly separates each object in discernible clusters.

\textbf{Ablation study:} To observe the contribution of the two main elements of RCFusion, multi-level feature extraction and recurrent fusion, we alternatively remove these elements and compare the performance with the full version of the method. Table \ref{tab:ocid} presents the results of these variations on OCID. It can be noticed that using only the features from the last layer of the RGB-/Depth-CNN (RCFusion - res5) drops the performance by $2\%$ in accuracy. This confirms that explicitly using features from several levels of abstraction improves the multi-modal recognition compared to only using the final features of single modalities. Analogously, substituting the RNN with a simple fully connected layer drops the performance by $3.1\%$ in accuracy. This confirms that a more sophisticated fusion mechanism that effectively combines the modalities while retaining the crucial information from the different levels of abstraction is crucial for obtaining a final discriminative RGB-D feature.

\setlength{\tabcolsep}{12pt}
\begin{table}[t]
\begin{center}
\caption{Accuracy (\%) of DECO~\cite{deco} and variations of RCFusion on Object Clutter Indoor Dataset \cite{ocid}. ``RCFusion - res5" refers to the the variation of RCFusion without extracting features from multiple layers, i.d. when only the features from the last residual layer (res5) are used for classification. ``RCFusion - fc" refers to the variaton of RCFusion with a fully connected layer used instead of the recurrent neural network for combining the RGB and depth features.}
\label{tab:ocid}
\begin{tabular}{lccc}
\hline\noalign{\smallskip}
\multicolumn{4}{c}{\head{Object Clutter Indoor Dataset}} \\
\noalign{\smallskip}
\cline{1-4}
\noalign{\smallskip}
Method & RGB & Depth & RGB-D \\
\hline\noalign{\smallskip}
 DECO~\cite{deco} &
 {\color{blue}\textbf{80.7}} & {\color{red}\textbf{36.8}} & 80.7 \\
 \noalign{\smallskip}
 \hline
 \noalign{\smallskip}
 RCFusion &
 {\color{red}\textbf{85.5}} & {\color{blue}\textbf{35.0}} & {\color{red}\textbf{91.6}} \\
  RCFusion - res5 &
 - & - & {\color{blue}\textbf{89.6}} \\
 RCFusion - fc &
 - & - & 88.5 \\
\noalign{\smallskip}
\hline
\end{tabular}
\end{center}
\end{table}
\setlength{\tabcolsep}{1.4pt}


\section{Discussion and Conclusion}
\label{sec:conclusion}
In this paper, we have presented RCFusion: a multi-modal deep neural network for RGB-D object recognition. Our method uses two streams of convolutional networks to extract RGB and depth features from multiple levels of abstraction. These features are then concatenated and sequentially fed to an RNN to obtain a compact RGB-D feature that is used by a softmax classifier for the final classification. We show the validity of our approach by outperforming the existing methods for RGB-D recognition on two standard benchmarks, RGB-D Object Dataset and JHUIT-50. In order to stress test our approach with some of the main challenges of robotic vision, we evaluate RCFusion on OCID. In fact, not only does this dataset present highly cluttered and occluded scenes, but it also provides few training samples. Despite these challenges, RCFusion presents compelling results on OCID and clearly marks the superiority of our multi-modal fusion mechanism. 

Our experiments also show that there is space for improvement. For the classes where the predictions of the RGB and the depth modality are often fooled by the same class, RCFusion can perform worse than the single modalities. In future work, we will investigate how to mitigate this problem by using an ensemble of different fusion mechanisms.

Due to their implementation-agnostic nature, the main concepts presented in this paper can be adapted to different tasks. The results obtained on object categorization encourage further research to extend this approach to higher level tasks, such as object detection and semantic segmentation. 

\balance




\section*{ACKNOWLEDGMENT}

This work has received funding from the European Union’s Horizon 2020 research and innovation program under the Marie Skłodowska-Curie grant agreement No. 676157, project ACROSSING, by the ERC grant 637076 - RoboExNovo (B.C.), and the CHIST-ERA project ALOOF (B.C.). The authors are also grateful to Tim Patten for the support in finalizing the work and Georg Halmetschlager for the help in the camera calibration process.




\end{document}